\documentclass{elsarticle}

\usepackage{amssymb}
\usepackage{amsthm}
\usepackage{amsmath}

\usepackage[colorlinks=true,linkcolor=blue,citecolor=blue,urlcolor=blue]{hyperref}

\journal{Arxiv}
\begin{document}
\begin{frontmatter}



\title{A Survey on Temporal Knowledge Graph: Representation Learning and Applications}


\author[a,b]{Li Cai}
\ead{lcai2020@stu.ecnu.edu.com}
\author[a]{Xin Mao}
\ead{xmao@stu.ecnu.edu.cn}
\author[a]{Yuhao Zhou}
\ead{51265900018@stu.ecnu.edu.cn}
\author[a]{Zhaoguang Long}
\ead{51265901014@stu.ecnu.edu.cn}
\author[c]{Changxu Wu}
\ead{wuchangxu@tsinghua.edu.cn} 
\author[a]{Man Lan\corref{cor1}}
\cortext[cor1]{Corresponding author.}
\ead{mlan@cs.ecnu.edu.cn}

\affiliation[a]{organization={School of Computer Science and Technology, East China Nomal University},
            addressline={3663 North Zhongshan Road}, 
            city={Shanghai},
            postcode={200062}, 
            country={China}}

\affiliation[b]{organization=
{College of Computer Science and Technology, Guizhou University},
            addressline={2708 South Huaxi Avenue}, 
            city={Guiyang},
            postcode={550025}, 
            state={Guizhou},
            country={China}}

\affiliation[c]{organization=
{Department of Industrial Engineering, Tsinghua University},
            addressline={30 Shuangqing Road}, 
            city={Beijing},
            postcode={100084}, 
            country={China}}
         
\begin{abstract}
Knowledge graphs have garnered significant research attention and are widely used to enhance downstream applications. However, most current studies mainly focus on static knowledge graphs, whose facts do not change with time, and disregard their dynamic evolution over time. As a result, temporal knowledge graphs have attracted more attention because a large amount of structured knowledge exists only within a specific period. Knowledge graph representation learning aims to learn low-dimensional vector embeddings for entities and relations in a knowledge graph. The representation learning of temporal knowledge graphs incorporates time information into the standard knowledge graph framework and can model the dynamics of entities and relations over time. In this paper, we conduct a comprehensive survey of temporal knowledge graph representation learning and its applications. We begin with an introduction to the definitions, datasets, and evaluation metrics for temporal knowledge graph representation learning. Next, we propose a taxonomy based on the core technologies of temporal knowledge graph representation learning methods, and provide an in-depth analysis of different methods in each category. Finally, we present various downstream applications related to the temporal knowledge graphs. In the end, we conclude the paper and have an outlook on the future research directions in this area.
\end{abstract}




\end{frontmatter}


\section{Introduction}

Knowledge graphs (KGs) describe the real world with structured facts. A fact consists of two entities and a relation connecting them, which can be formally represented as a triple \emph{(head, relation, tail)}, and an instance of a fact is \emph{(Barack Obama, make statement, Iran)}. Knowledge graph representation learning (KGRL) ~\citep{ji2021surveykg} seeks to learn the low-dimentional vector embeddings of entities and relations and use these embeddings for downstream tasks such as information retrieval~\citep{10.1145/3209978.3210187-InformationRetrieval}, question answering~\citep{10.1145/3437963.3441753-QuestionAnswering}, and recommender systems~\citep{10.1145/3460231.3474243-RecommenderSystem}. 

\begin{figure} [ht]
  \centering  \includegraphics[width=0.75\linewidth]{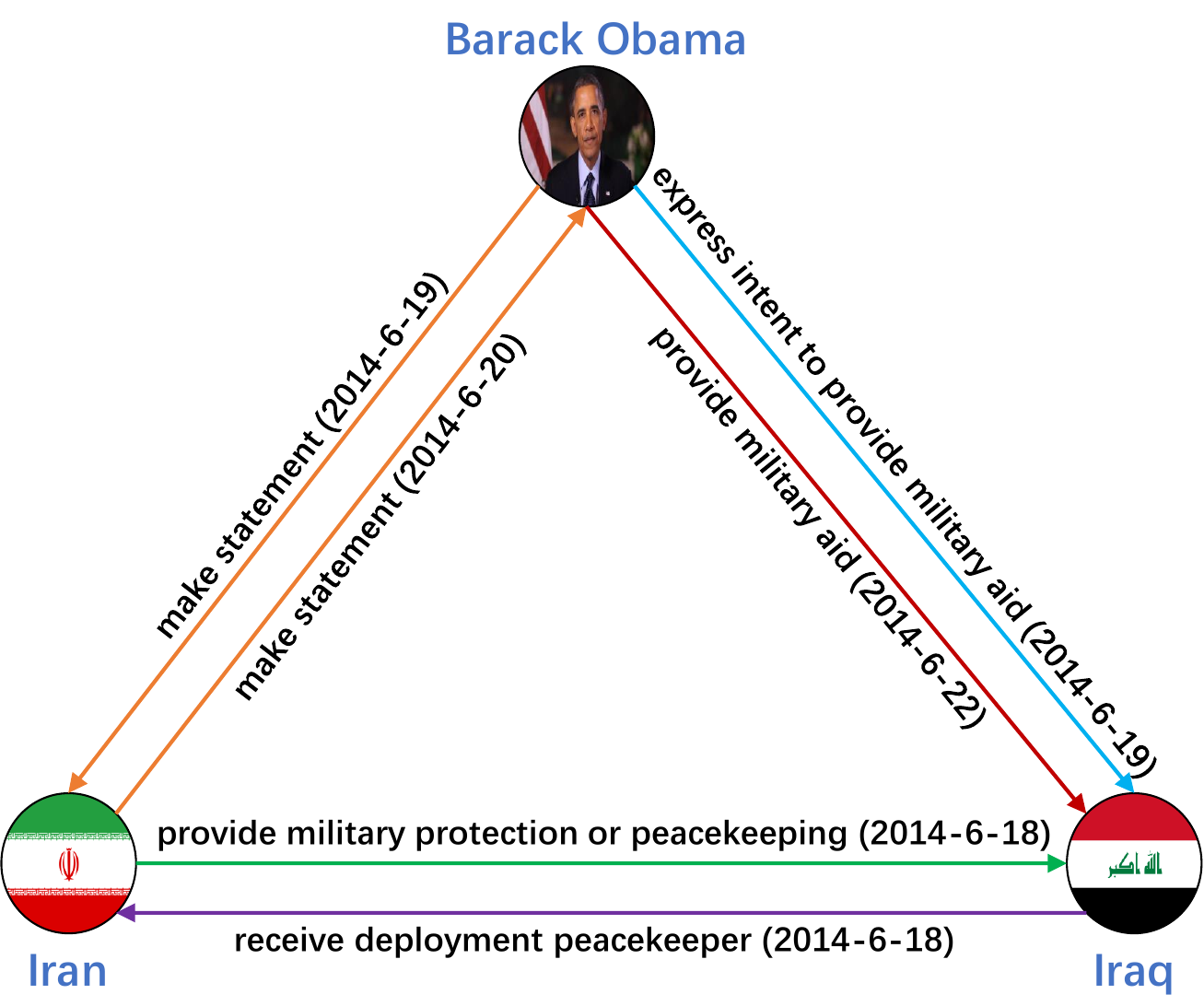}
  \caption{An example of temporal knowledge graph (a subgraph of ICEWS14).}
  \label{fig:sample}
\end{figure}

Existing KGs ignore the timestamp indicating when a fact occurred and cannot reflect their dynamic evolution over time. In order to represent KGs more accurately, Wikidata~\citep{10.1145/2629489-wikidata} and YOGO2~\citep{10.1145/1963192.1963296-yago2} add temporal information to the facts, and some event knowledge graphs~\citep{o2010crisis-icews,leetaru2013gdelt} also contain the timestamps indicating when the events occurred. The knowledge graphs with temporal information are called temporal knowledge graphs (TKGs). Figure~\ref{fig:sample} is a subgraph of the temporal knowledge graph. The fact in TKGs are expanded into quadruple \emph{(head, relation, tail, timestamp)}, a specific instance is \emph{(Barack Obama, make statement, Iran, 2014-6-19)}. 

The emergence of TKGs has led to increased researcher interest in temporal knowledge graph representation learning (TKGRL)~\citep{DBLP:conf/dsc/MoWJL21-SurveyTKG}. The acquired low-dimensional vector representations are capable of modelling the dynamics of entities and relations over time, thereby improving downstream applications such as time-aware knowledge reasoning~\citep{DBLP:conf/emnlp/JinQJR20-re-net}, entity alignment~\citep{xu-etal-2021-time-tea-gnn}, and question answering~\citep{DBLP:conf/acl/SaxenaCT20-cronkgqa}.

Temporal knowledge graph representation learning and applications are at the forefront of current research. Nevertheless, as of now, a comprehensive survey on the topic is not yet available. Ji et al.~\citep{ji2021surveykg} provides a survey on KGs, which includes a section on TKGs. However, this section only covers a limited number of early methods related to TKGRL. The paper~\citep{DBLP:conf/dsc/MoWJL21-SurveyTKG} is a survey on TKGs, with one section dedicated to introducing representation learning methods of static knowledge graphs, and only five models related to TKGRL are elaborated in detail. The survey~\citep{DBLP:journals/corr/abs-2201-08236-surveyTKGC} is about temporal knowledge graph completion (TKGC), and it focuses solely on the interpolation-based temporal knowledge graph reasoning application.

This paper comprehensively summarizes the current research of TKGRL and its related applications. Our main contributions are summarized as follows: (1) We conduct  an extensive investigation on various TKGRL methods up to the present, analyze their core technologies, and propose a new classification taxonomy. (2) We divide the TKGRL methods into ten distinct categories. Within each category, we provide detailed information on the key components of different methods and analyze the strengths and weaknesses of these methods. (3) We introduce the latest development of different applications related to TKGs, including temporal knowledge graph reasoning, entity alignment between temporal knowledge graphs, and question answering over temporal knowledge graphs. (4) We summarize the existing research of TKGRL and point out the future directions which can guide further work.

The remainder of this paper is organized as follows: Chapter 2 introduces the background of temporal knowledge graphs, including definitions, datasets, and evaluation metrics. 
Chapter 3 summarizes various temporal knowledge graph representation learning methods, including transformation-based methods, decomposition-based methods, graph neural networks-based methods, capsule network-based methods, autoregression-based methods, temporal point process-based methods, interpretability-based methods, language model methods, few-shot learning methods and others. 
Chapter 4 introduces the related applications of the temporal knowledge graph, such as temporal knowledge graph reasoning, entity alignment between temporal knowledge graphs, and question answering over temporal knowledge graphs. Chapter 5 highlights the future directions of Temporal Knowledge Graph Representation Learning (TKGRL), encompassing Scalability, Interpretability, Information Fusion, and the Integration of Large Language Models. Chapter 6 gives a conclusion of this paper.

\section{Background}

\subsection{Problem Formulation}
A temporal knowledge graph is a directed multi-relational graph containing structured facts. It is usually expressed as $G = (E, R, T, F)$, where $E$, $R$, and $T$ are the sets of entities, relations, and timestamps, respectively, and $F \subset E\times R\times E\times T$ is the set of all possible facts. A fact $f$ is denoted as $(h, r, t, \tau)$, where $h, r, t$, and $\tau$ are the head entity, relation, tail entity, and timestamp, respectively.

Take Figure~\ref{fig:sample} for example, where the entity set E contains \emph{(Barack Obama, Iran, Iraq)}, the relation set contains \emph{(make statement, express intent to provide military aid, provide military aid, provide military protection or peacekeeping, receive deployment peacekeeper)}, the time set contains \emph{(2014-6-18, 2014-6-19, 2014-6-20, 2014-6-22)}, and the fact set contains \emph{((Iran, Provide military protection or peacekeeping, Iraq, 2014-6-18)), (Iraq, receive deployment peacekeeper, Iran, 2014-6-18), (Barack Obama, Make statement, Iran, 2014-6-19), (Iran, Make statement, Barack Obama, 2014-6-20),(Barack Obama, express intent to provide military aid, Iraq, 2014-6-19), (Barack Obama, provide military aid, Iraq, 2014-6-22))}.

\begin{figure}
  \centering  \includegraphics[width=0.98\linewidth]{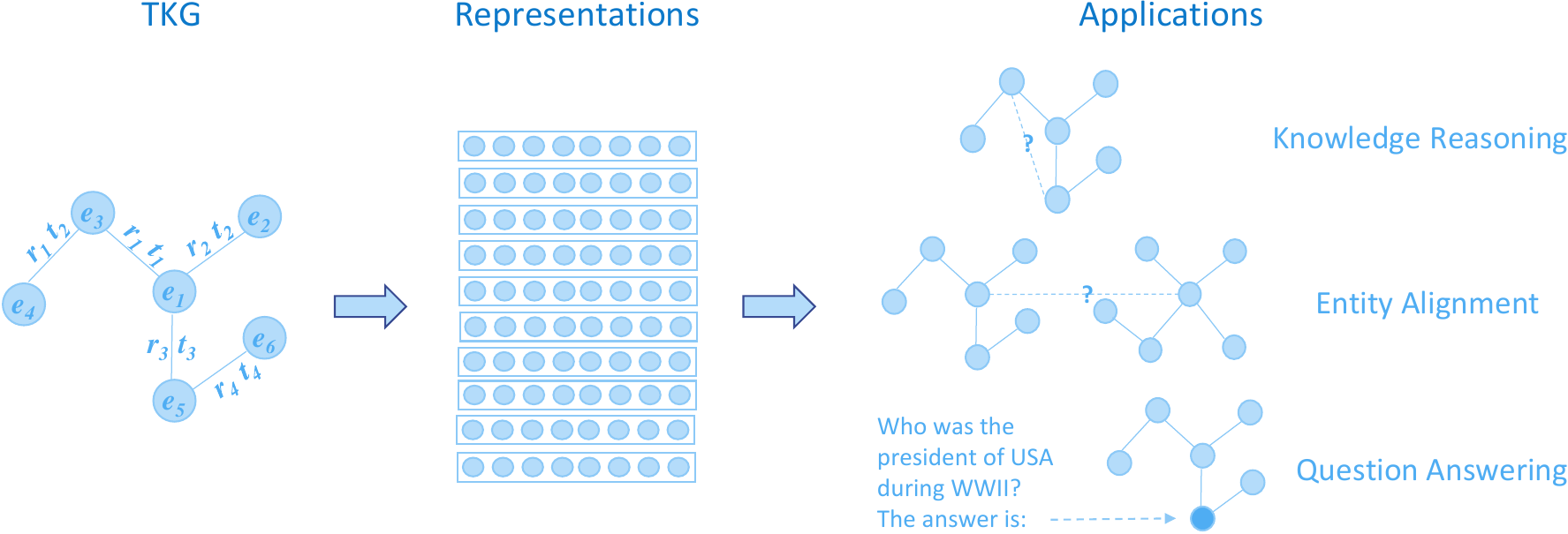}
  \caption{Temporal knowledge graph representation learning and applications.}
  \label{fig:applications}
\end{figure}

TKGRL aims to effectively learn low-dimensional vector representations of entities $\boldsymbol{h},\boldsymbol{t}$, relations $\boldsymbol{r}$, and timestamps $\boldsymbol{\tau}$ for downstream tasks such as knowledge reasoning, entity alignment and question answering (as shown in Figure~\ref{fig:applications}).

\subsection{Datasets}
There are four commonly used datasets for temporal knowledge graph representation learning.

\textbf{\emph{ICEWS}} 
The integrated crisis early warning system (ICEWS)~\citep{o2010crisis-icews} captures and processes millions of data points from digitized news, social media, and other sources to predict, track and respond to events around the world, primarily for early warning. Three subsets are typically used: ICEWS14, ICEWS05-15, and ICEWS18, which contain events in 2014, 2005-2015, and 2018, respectively.

\textbf{\emph{GDELT}} 
The global database of events, language, and tone (GDELT)~\citep{leetaru2013gdelt} is a global database of society. It includes the world's broadcast, print, and web news from across every country in over 100 languages and continually updates every 15 minutes.

\textbf{\emph{Wikidata}} 
The Wikidata~\citep{10.1145/2629489-wikidata} is a collaborative, multilingual auxiliary knowledge base hosted by the Wikimedia Foundation to support resource sharing and other Wikimedia projects. It is a free and open knowledge base that can be read and edited by both humans and machines. Many items in Wikidata have temporal information.

\textbf{\emph{YAGO}} 
The YAGO~\citep{10.1145/1963192.1963296-yago2} is a linked database developed by the Max Planck Institute in Germany. YAGO integrates data from Wikipedia, WordNet, and GeoNames. YAGO integrates WordNet's word definitions with Wikipedia's classification system, adding temporal and spatial information to many knowledge items.

\begin{table}
  \caption{Statistics of Datasets for Temporal Knowledge Graphs in different applications}
  \centering
  \label{tab:datasets}
  \scalebox{0.8}{
  \begin{tabular}{llrrrr}
    \hline
    Applications & Datasets & $|E|$ & $|R|$ & $|T|$ & $|F|$ \\
    \hline
    Knowledge Reasoning & ICEWS18 & 23,033 & 256 & 304 & 468,558\\
     & ICEWS14 & 7,128 & 230 & 365 & 90,730\\
 & ICEWS05-15 & 10,488 & 251 & 4,017 & 461,329\\
     & GDELT & 7,691 & 240 & 2,751 & 2,278,405\\
     & Wikidata & 12,554 & 24 & 232 & 669,934\\
     & YAGO & 10,623 & 10 & 189 & 201,089\\
  \hline
  Entity Alignment & DICEWS & 9,517/9,537 & 247/246 & 4,017 & 307,552/307,553\\
  & YAGO-WIKI & 49,626/49,222 & 11/30 & 245 & 221,050/317,814\\
  \hline
  Question Answering & ICEWS21 & - & 253 & 243 & -\\
  & Wikidata & 432,715 & 814 & 1,726 & 7,224,361\\
  \hline
\end{tabular}
}
\end{table}

The datasets of TKGs often require unique data processing methods for different downstream applications.  Table ~\ref{tab:datasets} presents the statistics of datasets for various tasks of TKGs. In knowledge reasoning tasks, datasets are typically divided into different training sets, validation sets, and test sets based on task type (interpolation and extrapolation). In entity alignment tasks, as the same entites in the real world need to be aligned between different KGs, a dataset always includes two temporal knowledge graphs that must be learned simultaneously. For question answering tasks, the datasets not only include temporal knowledge graphs used to search for answers but also include temporal-related questions (which have not been showed here).

\subsection{Evaluation Metrics}
The evaluation metrics for verifying the performance of TKGRL are $MRR$ (mean reciprocal rank) and $Hit@k$.

$\boldsymbol{MRR}$ The $MRR$ represents the average of the reciprocal ranks of the correct answers. It can be calculated as follows: 
\begin{equation}
MRR = \frac{1}{|S|}\sum_{i=1}^{|S|} \frac{1}{rank_i}
\label{eq1}
\end{equation}
where $S$ is the set of all correct answers, $|S|$ is the number of the sets. The predicted result is a set sorted by the probability of the answer from high to low, and $rank_i$ is the rank of the \emph{i-th} correct answer in the prediction result. The higher the MRR, the better the performance.

$\boldsymbol{Hit@k}$ The $Hits@k$ reports the proportion of correct answers in the top $k$ predict results. It can be calculated by the following equation: 
\begin{equation}
Hit@k = \frac{1}{|S|}\sum_{i=1}^{|S|} \boldsymbol{I}(rank_i\le{k})
\label{eq2}
\end{equation}
where the $|S|$ and $rank_i$ are the same as above, $\boldsymbol{I}(\cdot)$ is the indicator function (If the condition ($rank_i\le{k}$) is true, the function value is 1, otherwise 0). Typically, $k$ is $1,3,5$ or $10$. $Hit@1$ represents the percentage that the first-rank predicted result is the correct answer, equivalent to the $Accuracy$. $Hit@10$ represents the percentage of the top ten predictions containing correct answers. The higher the $Hit@k$, the better the performance.
\begin{figure}
  \centering  \includegraphics[width=0.99\linewidth]{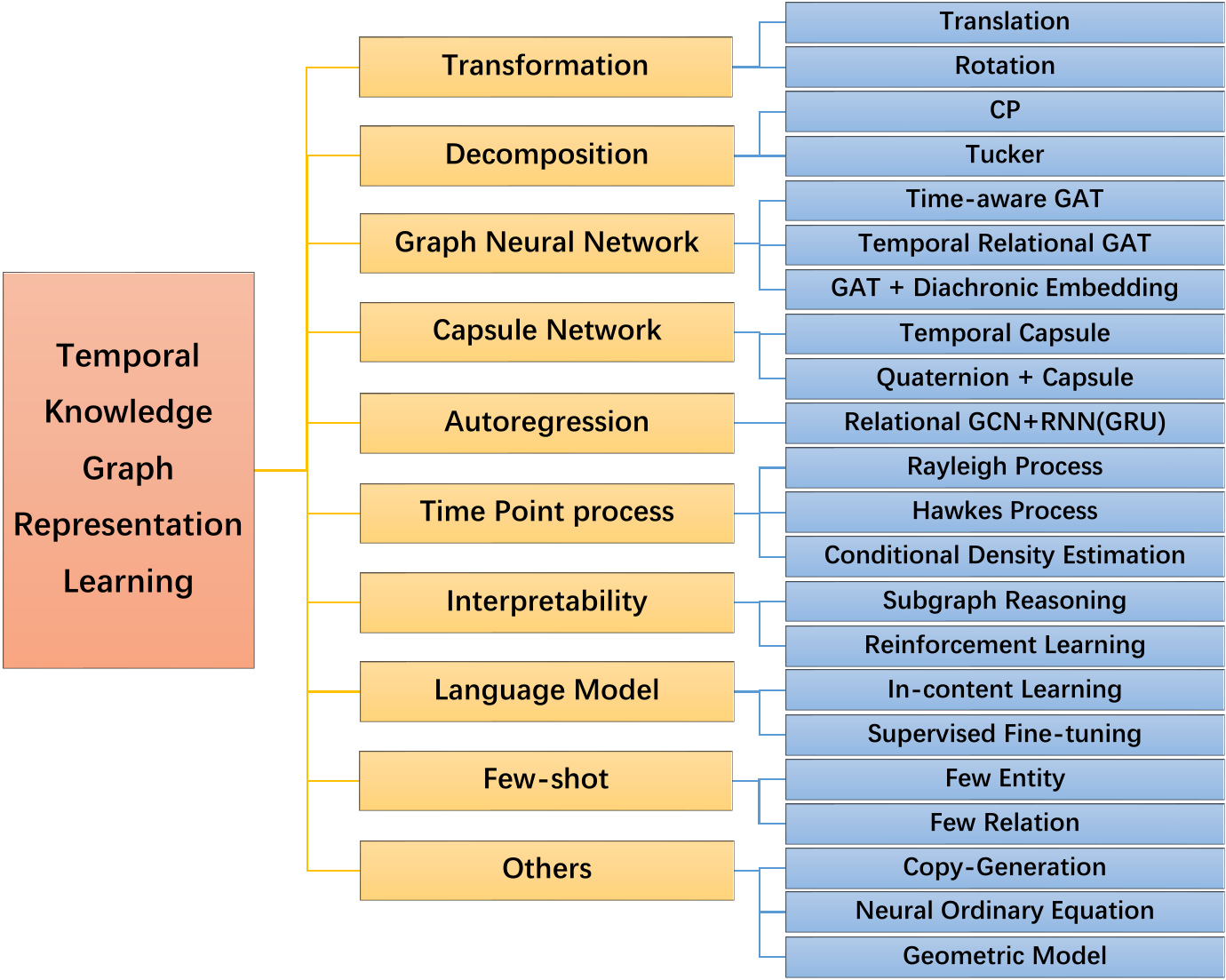}
  \caption{The categorization of temporal knowledge graph representation learning methods.}
  \label{fig:categories}
\end{figure}
\section{Temporal Knowledge Graph Representation Learning Methods}

Compared to KGs, TKGs contain additional timestamps, which are taken into account in the construction of TKGRL methods. These methods can be broadly categorized into transformation-based, decomposition-based, graph neural networks-based, and capsule network-based approaches. Additionally, temporal knowledge graphs can be viewed as sequences of snapshots captured at different timestamps or events that occur over continuous time, and can be learned using autoregressive and temporal point process techniques. Moreover, some methods prioritize interpretability, language model, and few-shot learning. Thus, based on the core technologies employed by TKGRL methods, we group them into the following categories: transformation-based methods, decomposition-based methods, graph neural networks-based methods, capsule network-based methods, autoregression-based methods, temporal point process-based methods, interpretability-based methods, language model methods, few-shot learning methods, and others. A visualization of the categorization of TKGRL methods is presented in Figure~\ref{fig:categories}.

The notations in these methods are varied, and we define our notations to describe them uniformly. We use lower-case letters to denote scalars, bold lower-case letters to denote vectors, bold upper-case letters to denote matrices, bold calligraphy upper-case letters to denote order 3 tensors, and bold script upper-case letters to denote order 4 tensors. The main notations and their descriptions are listed in table~\ref{tab:notations}.

\begin{table}
  \caption{Notations and descriptions}
  \centering
  \label{tab:notations}
  \scalebox{0.9}{
  \begin{tabular}{ll}
    \hline
    Notation & Description \\
    \hline
    $E, R, T, F$ & The sets of entities, relations, timestamps, and facts\\
    $(h,r,t,\tau)$ & A quadruple of head entity, relation, tail entity and timestamps\\
    $\boldsymbol{h},\boldsymbol{r},\boldsymbol{t},\boldsymbol{\tau}$ & Embedding of head entity, relation, tail entity and timestamps\\
    $\boldsymbol{h_{\tau}},\boldsymbol{r_{\tau}},\boldsymbol{t_{\tau}}$ & Temporal embedding of head entity, relation, and tail entity\\
    $\boldsymbol{{\tau}_e},\boldsymbol{{\tau}_r}$ & Entity and relation embedding of time\\
    $\mathbb{R,C,H,G}$ & Set of real, complex, hypercomplex and multivectors\\
    $\Vert \cdot \Vert_{1/2}$ & L1-norm or L2-norm\\
    $d(\boldsymbol{a,b})$ & Distance between $\boldsymbol{a}$ and $\boldsymbol{b}$\\
    $\boldsymbol{w},\boldsymbol{W},\boldsymbol{\mathcal{W}}$ & Parameters in vector, matrix, and order 3 tensor form \\    
    $\circ$ & Hadamard (element-wise) product\\
    $\otimes$ & Kronecker product\\
    $\odot$ & Hamilton product\\
    $\left\langle \cdot \right\rangle$ & Dot product\\  
    $re(\boldsymbol{t})$ & Real vector component of $\boldsymbol{t}$\\ 
    $\overline{\boldsymbol{t}}$ & Conjugate vector component of $\boldsymbol{t}$ \\
    $\boldsymbol{r}^{-1}$ & Inverse of vector $\boldsymbol{r}$ \\
    $\boldsymbol{w}^{T}$ & Transposition of vector $\boldsymbol{w}$ \\
  \hline
\end{tabular}
}
\end{table}

\subsection{Transformation-based Methods}
In the transformation-based method, timestamps or relations are regarded as the transformation between entities. The representation learning of TKGs is carried out by integrating temporal information or mapping entities and relations to temporal hyperplanes based on the existing KGRL methods. There are translation-based transformations and rotation-based transformations.

\textbf{Translation-based}
The translation-based representation learning methods of TKGs are developed based on TransE [11]. TransE is a classic translation-based KGRL method. It regards the relation $r$ as a translation from the head entity $h$ to the tail entity $t$, that is, $\boldsymbol{h}+\boldsymbol{r}\approx \boldsymbol{t}$, so the score function is $\Vert\boldsymbol{h}+\boldsymbol{r}-\boldsymbol{t}\Vert_{1/2}$.

TTransE~\citep{10.1145/3184558.3191639-TTransE}
concatenate the temporal information to the relations, and the quadruple $(h, r, t, \tau)$ is converted into a triple $(h, [r|\tau], t)$. The new relation embeddings are expressed as $\boldsymbol{r}+\boldsymbol{\tau}$, and the score function is $\Vert\boldsymbol{h}+\boldsymbol{r}+\boldsymbol{\tau}-\boldsymbol{t}\Vert_{1/2}$.

Compared with the simple connection of temporal information and relations by TTransE, TA-TransE~\citep{Garcia-DuranDN18tatranse} creates more expressive synthetic relations, which treats relations as sequences containing temporal information, such as \emph{(make statement,2y,0y,1y,4y,06m,1d,9d)}. Then TA-TransE use LSTM~\citep{DBLP:journals/neco/HochreiterS97-lstm} to learn the relation embedding $\boldsymbol{r}_{seq}$, and the score function is $\Vert\boldsymbol{h}+\boldsymbol{r}_{seq}-\boldsymbol{t}\Vert_{2}$.

Different from TTransE and TA-TransE, HyTE~\citep{dasgupta18hyte} learns the representations of entities, relations, and temporal information jointly. It splits the temporal knowledge graph into multiple static subgraphs, each of which corresponds to a timestamp. Then it incorporates time in the entity-relation space by associating each timestamp with a corresponding hyperplane. For a timestamp $\tau$, the corresponding hyperplane is $\boldsymbol{w}_{\tau}$, if a triple $(h, r, t)$ is valid for the timestamp, then the projected representations of the triple on the hyperplane are    
$\boldsymbol{h}_{\tau} = \boldsymbol{h}-(\boldsymbol{w}_{\tau}^T\boldsymbol{h})\boldsymbol{w}_{\tau},
\boldsymbol{r}_{\tau} = \boldsymbol{r}-(\boldsymbol{w}_{\tau}^T\boldsymbol{r})\boldsymbol{w}_{\tau},
\boldsymbol{t}_{\tau} = \boldsymbol{t}-(\boldsymbol{w}_{\tau}^T\boldsymbol{t})\boldsymbol{w}_{\tau}$, 
the score function is $\Vert \boldsymbol{h}_{\tau} + \boldsymbol{r}_{\tau} -\boldsymbol{t}_{\tau} \Vert_{1/2}$. 

\textbf{Rotation-based} 
The translation-based methods can infer the inverse and composition relation patterns, except the symmetry pattern, because all the symmetry relations will be presented by a 0 vector, so the rotation-based transformation methods appear. RotatE~\citep{DBLP:conf/iclr/SunDNT19-rotate} is the first method based on rotation. It regards relation as a rotation from head entity to tail entity, expands the representation space from real-valued point-wise space ($\mathbb{R}^d$) to complex vector space ($\mathbb{C}^d, \ \boldsymbol{c = a+bi}, \ \boldsymbol{a,b} \in \mathbb{R}^d $), and expects $\boldsymbol{t} = \boldsymbol{h} \circ \boldsymbol{r}$, where $\circ$ is Hadamard (element-wise) product. The score function is $ \Vert \boldsymbol{h} \circ \boldsymbol{r}-\boldsymbol{t} \Vert_1$.

Tero~\citep{xu20tero} regards timestamps as the rotation of entities in complex space. The mapping function is $\boldsymbol{h}_{\tau} = \boldsymbol{h} \circ \boldsymbol{\tau}, \boldsymbol{t}_{\tau} = \boldsymbol{t} \circ \boldsymbol{\tau}$, then the entity representations contain the temporal information. It regards the relation embedding $\boldsymbol{r}$ as the translation from the time-specific head entity embedding $\boldsymbol{h}_{\tau}$ to the conjugate of the time-specific tail entity embedding $\overline{\boldsymbol{t}}_{\tau}$. The score function is $\Vert \boldsymbol{h}_{\tau}+\boldsymbol{r}-\overline{\boldsymbol{t}}_{\tau} \Vert_1$. 

ChronoR~\citep{sadeghian2021chronor} proposes a model with a k-dimensional rotation transformation. It regards the relations and timestamps as the rotation of the head entity in k-dimensional space and expects the head entity falls near the tail entity after the rotation. The rotation is defined as $\boldsymbol{h}_{r,\tau} = \boldsymbol{h} \circ \left[\boldsymbol{r}|\boldsymbol{\tau}\right] \circ \boldsymbol{r}_2$, where $|$ presents concatenation operator, $\boldsymbol{r}_2$ is the static relation representation without considering temporal information. The score function is $\left\langle \boldsymbol{h}_{r,\tau}, \boldsymbol{t}\right\rangle$.       

RotateQVS~\citep{chen-etal-2022-rotateqvs} utilizes a hypercomplex (quaternion) vector space ($ \ \mathbb{H}^d, \\
\boldsymbol{q = a+bi+cj+dk}, \ \boldsymbol{a,b,c,d} \in \mathbb{R}^d $) to represent entities, relations, and timestamps. The time-specific entity representations are learned through temporal rotation transformations in 3D space, where the mapping function is denoted as $\boldsymbol{h}_{\tau} = \boldsymbol{\tau} \boldsymbol{h} \boldsymbol{\tau}^{-1}, \boldsymbol{t}_{\tau} = \boldsymbol{\tau} \boldsymbol{t} \boldsymbol{\tau}^{-1} $, and the score function is defined as $\Vert \boldsymbol{h}_{\tau}+\boldsymbol{r}-\overline{\boldsymbol{t}}_{\tau} \Vert_2$.

\subsection{Decomposition-based Methods}
The main task of representation learning is to learn the low-dimensional vector representation of the knowledge graph. Tensor decomposition has three applications: dimension reduction, missing data completion, and implicit relation mining, which meet the needs of knowledge graph representation learning. The knowledge graph consists of triples and can be represented by an order 3 tensor. For the temporal knowledge graph, the additional temporal information can be represented by an order 4 tensor, and each tensor dimension is the head entity, relation, tail entity, and timestamp, respectively. Tensor decomposition includes Canonical Polyadic (CP) decomposition~\citep{hitchcock1927expression-cp} and Tucker decomposition~\citep{tucker1966some}.

\textbf{CP decomposition}
For an order 3 tensor $\boldsymbol{\mathcal{X}} \in \mathbb{R}^{n_1 \times n_2 \times n_3}$, CP decomposition factorize $\boldsymbol{\mathcal{X}}$ as $\boldsymbol{\mathcal{X}} \approx \left\langle \boldsymbol{A},\boldsymbol{B},\boldsymbol{C} \right\rangle = \sum^d_{\alpha=1} \boldsymbol{A}_{:,\alpha} \otimes \boldsymbol{B}_{:,\alpha} \otimes \boldsymbol{C}_{:,\alpha}$, where $\otimes$ presents the Kronecker product, $ \boldsymbol{A} \in \mathbb{R}^{n_1 \times d}, \boldsymbol{B} \in \mathbb{R}^{n_2 \times d}, \boldsymbol{C} \in \mathbb{R}^{n_3 \times d} $ are decomposed matrices.

DE-SimplE~\citep{GoelKBP20De-SimplE} learns the diachronic embeddings of entities and uses the score function in SimplE~\citep{NEURIPS2018_b2ab0019-SimplE} for temporal knowledge graph completion. SimplE is an enhancement of CP decomposition, which can learn the entity embeddings independently. The diachronic entity embedding is defined as $\boldsymbol{e}_{\tau}[n]=\left\{\begin{array}{ll} \boldsymbol{e}[n]\sigma(\boldsymbol{w}[n]\tau+\boldsymbol{b}[n]),  & 1 \leq n \leq \gamma d  \\ \boldsymbol{e}[n],  & \gamma d \le n \leq d\end{array}\right.$, and the score function is $\frac{1}{2}(\left\langle \boldsymbol{h}_{\tau},\boldsymbol{r},\boldsymbol{t}_{\tau} \right\rangle + \left\langle \boldsymbol{t}_{\tau},\boldsymbol{r^{-1}},\boldsymbol{h}_{\tau} \right\rangle)$. 

T-SimplE~\citep{LinS20T-SimplE} regards the temporal knowledge graph as an order 4 tensor and uses tensor decomposition to learn the embeddings of entities, relations, and timestamps. The score function is $\frac{1}{2}(\left\langle \boldsymbol{h},\boldsymbol{r},\boldsymbol{t},\boldsymbol{\tau} \right\rangle + \left\langle \boldsymbol{t},\boldsymbol{r}^{-1},\boldsymbol{h},\boldsymbol{\tau} \right\rangle)$.

TComplEx~\citep{LacroixOU20TComplEx} extends the representation space to complex vector space and uses the tensor decomposition based on ComplEx~\citep{trouillon2016complex}. ComplEx is a simple link prediction method with complex embeddings. TComplEx adds the temporal embeddings and takes $re(\left\langle \boldsymbol{h},\boldsymbol{r},\overline{\boldsymbol{t}},\boldsymbol{\tau} \right\rangle)$ as the score function.

TLT-KGE~\citep{10.1145/3511808.3557233-TLTKGE} integrates time representation into the representation of entities and relations, and verifies its validity in both complex and hypercomplex (quaternion) spaces. Specifically, in the case of complex space, the representation of entities and relations is a combination of their semantic representation (real part) and temporal representation (imaginary part): $\boldsymbol{h}^c_{\tau}=\boldsymbol{h}+\boldsymbol{\tau_ei}$, $\boldsymbol{t}^c_{\tau}=\boldsymbol{t}+\boldsymbol{\tau_ei}$, $\boldsymbol{r}^c_{\tau}=\boldsymbol{r}+\boldsymbol{\tau_ri}$. Then calculate $\boldsymbol{C}(h,r,\tau)=\boldsymbol{h}^c_{\tau} \odot \boldsymbol{r}^c_{\tau}=(\boldsymbol{h} \circ \boldsymbol{r} - \boldsymbol{\tau_e} \circ \boldsymbol{\tau_r}) + (\boldsymbol{h} \circ \boldsymbol{\tau_r} + \boldsymbol{\tau_e} \circ \boldsymbol{r})\boldsymbol{i} = \boldsymbol{c_0} + \boldsymbol{c_1i}$, let $\boldsymbol{C'}(h,r,\tau) = \boldsymbol{c_1} + \boldsymbol{c_0i}$, the score function is $\phi^c(h,r,t,\tau)=\left<\boldsymbol{C}(h,r,\tau),\boldsymbol{t^c_{\tau}}\right> + \left<\boldsymbol{C'}(h,r,\tau),\boldsymbol{t^c_{\tau}}\right> = \left<\boldsymbol{c_0},\boldsymbol{t}\right> + \left<\boldsymbol{c_1},\boldsymbol{\tau_e}\right> + \left<\boldsymbol{c_0},\boldsymbol{\tau_e}\right> + \left<\boldsymbol{c_1},\boldsymbol{t}\right>$.
In the case of hypercomplex space, half of the quaternion is dedicated to semantic representation while the other half is dedicated to temporal representation:  
$\boldsymbol{h}^q_{\tau}=\boldsymbol{h_a}+\boldsymbol{h_bi}+\boldsymbol{\tau_{e,c}j}+\boldsymbol{\tau_{e,d}k}$, 
$\boldsymbol{t}^q_{\tau}=\boldsymbol{t_a}+\boldsymbol{t_bi}+\boldsymbol{\tau_{e,c}j}+\boldsymbol{\tau_{e,d}k}$,
$\boldsymbol{r}^q_{\tau}=\boldsymbol{r_a}+\boldsymbol{r_bi}+\boldsymbol{\tau_{r,c}j}+\boldsymbol{\tau_{r,d}k}$, the score function is $\phi^q(h,r,t,\tau)=\left<\boldsymbol{Q}(h,r,\tau),\boldsymbol{t^q_{\tau}}\right> + \left<\boldsymbol{Q'}(h,r,\tau),\boldsymbol{t^q_{\tau}}\right>$.
In addition, the author introduced a shared time window module for capturing the periodicity of entities and relationships, as well as a relation-timestamp composition module to model relation representations at specific time. The model was further regularized with respect to entities, relationships, and time to improve its generalization performance and mitigate overfitting. Experimental results indicate that the model in hypercomplex space achieves state-of-the-art results, but it also requires a large amount of storage space.

\textbf{Tucker decomposition}
Tucker decomposition is a more general tensor decomposition technique, and CP decomposition is its particular case. Tucker decomposition factorizes a tensor into a core tensor multiplied by a matrix in each dimension, such as $\boldsymbol{\mathcal{X}} \approx \left\langle \boldsymbol{\mathcal{W}};\boldsymbol{A},\boldsymbol{B},\boldsymbol{C} \right\rangle = \left\langle \boldsymbol{\mathcal{W}}_{\times 1} \boldsymbol{A}_{\times 2} \boldsymbol{B}_{\times 3} \boldsymbol{C} \right\rangle$, where $\boldsymbol{\mathcal{W}} \in \mathbb{R}^{d_1 \times d_2 \times d_3}$ is the core tensor, $ \boldsymbol{A} \in \mathbb{R}^{n_1 \times d_1}, \boldsymbol{B} \in \mathbb{R}^{n_2 \times d_2}, \boldsymbol{C} \in \mathbb{R}^{n_3 \times d_3} $ are decomposed matrices. When $\boldsymbol{\mathcal{W}}$ is a hyper-diagonal tensor and $d_1=d_2=d_3$, Tucker decomposition is equivalent to CP decomposition. TuckER~\citep{balazevic-etal-2019-tucker} introduces the Tucker decomposition for link prediction on knowledge graphs, which regards the knowledge graph as an order 3 tensor and decomposes it into a core tensor, head entity embedding, relation embedding, and tail entity embedding, the score function is $\left\langle \boldsymbol{\mathcal{W}}; \boldsymbol{h},\boldsymbol{r},\boldsymbol{t} \right\rangle$. 

TuckERT~\citep{Shao22TuckERT} proposes an order 4 tensor decomposition model based on Tucker decomposition. The model is fully expressive and effective for temporal knowledge graph completion. The score function is $\left\langle \boldsymbol{\mathcal{M}}; \boldsymbol{h},\boldsymbol{r},\boldsymbol{t}, \boldsymbol{\tau} \right\rangle$, where $\boldsymbol{\mathcal{M}}$ is an order 4 tensor.

\subsection{Graph Neural Networks-based Methods} \label{sec:gnn-based}
Graph Neural Networks (GNN)~\citep{4700287-gnn} have powerful structure modeling ability. The entity can enrich its representation with the attribute feature and the global structure feature by GNN. Typical graph neural networks include Graph Convolutional Networks (GCN)~\citep{DBLP:conf/iclr/KipfW17-gcn} and Graph Attention Networks (GAT)~\citep{DBLP:conf/iclr/VelickovicCCRLB18-gat}. GCN gets the representation of nodes by aggregating neighbor embeddings, and GAT uses a multi-head attention mechanism to get the representation of nodes by aggregating weighted neighbor embedding. The knowledge graph is a kind of graph that has different relations. The relation-aware graph neural networks are developed to learn the representations of entities in the knowledge graph. Relational Graph Convolutional Networks (R-GCN)~\citep{DBLP:conf/esws/SchlichtkrullKB18-r-gcn} is a graph neural network model for relational data. It learns the representation for each relation and obtains entity representation by aggregating neighborhood information under different relation representations. Temporal knowledge graphs have additional temporal information, and some methods enhance the representation of entities by a time-aware mechanism. 

TEA-GNN~\citep{xu-etal-2021-time-tea-gnn} learns entity representations through a time-aware graph attention network, which incorporates relational and temporal information into the GNN structure. Specifically, it assigns different weights to different entities with orthogonal transformation matrices computed from the neighborhood's relational embeddings and temporal embeddings and obtains the entity representations by aggregating the neighborhood.

TREA~\citep{10.1145/3485447.3511922-TREA} learns more expressive entity representation through a temporal relational graph attention mechanism. It first maps entities, relations, and timestamps into an embedding space, then integrates entities' relational and temporal features through a temporal relational graph attention mechanism from their neighborhood, and finally, uses a margin-based log-loss to train the model and obtains the optimized representations.
 
DEGAT~\citep{10.1007/978-3-031-10983-6_55-zhu22DEGAT} proposes a dynamic embedding graph attention network. It first uses the GAT to learn the static representations of entities by aggregating the features of neighbor nodes and relations, then adopts a diachronic embedding function to learn the dynamic representations of entities, and finally concatenates the two representations and uses the ConvKB as the decoder to obtain the score.

$T^2$TKG\citep{zhang2023learning}, or Latent relations Learning method for Temporal Knowledge Graph reasoning, is a novel approach that addresses the limitations of existing methods in explicitly capturing intra-time and inter-time latent relations for accurate prediction of future facts in Temporal Knowledge Graphs. It first employs a Structural Encoder (SE) to capture representations of entities at each timestamp, encoding their structural information. Then, it introduces a Latent Relations Learning (LRL) module to mine and exploit latent relations both within the same timestamp (intra-time) and across different timestamps (inter-time). Finally, the method fuses the temporal representations obtained from SE and LRL to enhance entity prediction tasks.

\subsection{Capsule Network-based Methods}
CapsNet is first proposed for computer vision tasks to solve the problem that CNN needs lots of training data and cannot recognize the spatial transformation of targets. The capsule network is composed of multiple capsules, and one capsule is a group of neurons. The capsules in the lowest layer are called primary capsules, usually implemented by convolution layers to detect the presence and pose of a particular pattern (such as eyes, nose, or mouth). The capsules in the higher level are called routing capsules, which are used to detect more complex patterns (such as faces). The output of a capsule is a vector whose length represents the probability that the pattern is present and whose orientation represents the pose of the pattern. 

CapsE~\citep{nguyen-etal-2019-CapsE} explores the application of capsule network in the knowledge graph. It represents a triplet as a three-column matrix in which each column represents the embedding of the head entity, relation, and tail entity, respectively. The matrix was fed to the capsule network, which first maps the different features of the triplet by a CNN layer, then captures the various patterns by the primary capsule layer, and finally routes the patterns to the next capsule layer to obtain the continuous output vector whose length indicates whether the triplet is valid.

TempCaps~\citep{fu-etal-2022-tempcaps} incorporates temporal information and proposes a capsule network for temporal knowledge graph completion. The model first selects the neighbors of the head entity in a time window and obtains the embeddings of these neighbors with the capsules, then adopts a dynamic routing process to connect the lower capsules and higher capsules and gets the head entity embedding, and finally uses a multi-layer perceptron (MLP)~\citep{gardner1998artificial-mlp} as the decoder to produce the scores of all candidate entities.

BiQCap~\citep{DBLP:conf/dasfaa/ZhangLLFZW23-BiQCap} utilizes biquaternions in hypercomplex space ($\mathbb{H}^d, \boldsymbol{q = a+bi} \\ \boldsymbol{+cj+dk}, \ \boldsymbol{a,b,c,d} \in \mathbb{C}^d$) and capsule networks to learn the representations of entities, relations, and timestamps. The model first represents the head entities, relations, tail entities, and timestamps using biquaternions. It then treats the timestamp as the translation of the head entity and obtains the time-specific representation ($\boldsymbol{h_\tau = h + \tau}$) of the head entity. Next, it rotates $\boldsymbol{h_\tau}$ and considers the rotated representation ($\boldsymbol{h_{\tau,r}}=\boldsymbol{h_\tau} \odot \boldsymbol{r}$) as being close to the tail entity to train the model. The score function is $\Vert \boldsymbol{h}_{\tau,r} \circ \boldsymbol{b} -\boldsymbol{t} \Vert$, where $\boldsymbol{b}$ is a learnable parameter. Finally, the trained representations of entities, relations, and timestamps are fed into the capsule network to obtain the final representation.

DuCape~\citep{10.1145/3589644-DuCape} represents entities, relations, and time using dual quaternions in hypercomplex space ($\mathbb{H}^d, \ \boldsymbol{q = q_0 + q_1 \xi}, \ \boldsymbol{q_0 = a_0+} \\ \boldsymbol{b_0i+c_0j+d_0k}, \ \boldsymbol{q_1 = a_1+b_1i+c_1j+d_1k}, \ \boldsymbol{a_0,\,b_0,\,c_0,\,d_0,\,a_1,\,b_1,\,c_1,} \\ \boldsymbol{d_1} \in \mathbb{R}^d$). Dual quaternions enable modeling of both rotation and translation operations simultaneously. The model first transforms the head entity through relations and timestamps in the dual quaternion space, where the representation is close to that of the tail entity. The scoring function is $\Vert \boldsymbol{h} \odot \boldsymbol{r} \odot \boldsymbol{\tau} -\boldsymbol{t} \Vert$. The learned representations are then inputted into the capsule network to obtain the final representation.

\subsection{Autoregression-based Methods}
The representation learning methods based on autoregression consider that the above methods cannot model the evolution of the temporal knowledge graph over time, so they cannot predict the knowledge graph in the future. It assumes that the knowledge graph of time $\tau$ can be inferred from the knowledge graph of last time and samples the temporal knowledge graph $G$ according to the timestamp to obtain a series of subgraphs (or snapshots) $\{ G_1, G_2,..., G_T \}$. Each subgraph contains the facts of the TKG at a timestamp. By modeling the subgraphs recurrently, the autoregression-based methods learn the evolutional representations of entities and relations to infer the facts $G_{T+1}$ in the future. 

RE-NET~\citep{DBLP:conf/emnlp/JinQJR20-re-net} proposes a recurrent event network to model the TKG for predicting future facts. It believes that the facts in $G_{T}$ at timestamp $T$ depend on the facts in the past m subgraphs $G_{T-1:T-m}$ before $T$. It first uses the R-GCN to learn the global structural representations and local neighborhood representations of the head entity at each timestamp. Then it utilizes the gated recurrent units (GRU)~\citep{DBLP:conf/emnlp/ChoMGBBSB14-gru} to update the above representations and pass these representations to an MLP as a decoder to infer the facts at timestamp $T$. This method only models the representations of the specific entity and relation in the query triple, ignoring the structural dependency between all triples in each subgraph, which may lose some important information from the entities not in the query triple.

Glean~\citep{10.1145/3394486.3403209-glean} thinks that most of the existing representation learning methods use the structural information of TKG, ignoring the unstructured information such as semantic information of words. It proposes a temporal graph neural network with heterogeneous data fusion. Specifically, it first constructs temporal event graphs based on historical facts at each timestamp and temporal word graphs from event summaries at each timestamp. Then it uses the CompGCN~\citep{DBLP:conf/iclr/VashishthSNT20-compgcn} to learn the structural representations of entities and relations in the temporal event graphs and the GCN to learn the textual semantic representations of entities and relations in the temporal word graphs. Finally, it fuses the two representations and utilizes a recurrent encoder to model temporal features for final prediction. 

RE-GCN~\citep{10.1145/3404835.3462963-re-gcn} splits the TKG into a sequence of KG according to the timestamps and encodes the facts in the past m steps recurrently to predict the entities and relations in the future. This model proposes a recurrent evolution network based on a graph convolution network to model the evolutional representations by incorporating the structural dependencies among concurrent facts, the sequential patterns across temporally adjacent facts, and the static properties. It uses the relation-aware GCN to capture the structural dependency and utilizes GRU to obtain the sequential pattern. Then it combines the static properties learned by R-GCN as a constraint to learn the evolutional representations of entities and relations and adopts ConvTransE~\citep{DBLP:conf/aaai/ShangTHBHZ19-convtranse} as the decoder to predict the probability of entities and relations at next timestamp.

TiRGN~\citep{DBLP:conf/ijcai/LiS022-tirgn} argues that the above methods can only capture the local historical dependence of the adjacent timestamps and cannot fully learn the historical characteristics of the facts. It proposes a time-guided recurrent graph neural network with local-global historical patterns which can model the historical dependency of events at adjacent snapshots with a local recurrent encoder, the same as RE-GCN, and collect repeated historical facts by a global history encoder. The final representations are fed into a time-guided decoder named Time-ConvTransE/Time-ConvTransR to predict the entities and relations in the future. 

Cen~\citep{DBLP:conf/acl/LiGJPL000GC22-cen} believes that modeling historical facts with fixed time steps could not discover the complex evolutional patterns that vary in length. It proposes a complex evolutional network that use the evolution unit in RE-GCN as a sequence encoder to learn the representations of entities in each subgraph and utilizes the CNN as the decoder to obtain the feature maps of historical snapshots with different length. The curriculum learning strategy is used to learn the complex evolution pattern with different lengths of historical facts from short to long and automatically select the optimal maximum length to promote the prediction.

\subsection{Temporal Point Process-based Methods}
The autoregression-based methods sample the TKG into discrete snapshots according to a fixed time interval, which cannot effectively model the facts with irregular time intervals. Temporal point process (TPP)~\citep{daley2008introduction-tpp} is a stochastic process composed of a series of events in a continuous time domain. The representation learning methods based on TPP regard the TKG as a list of events changing continuously with time and formalize it as $(G, O)$, where $G$ is the initialized TKG at time $\tau_0$, $O$ is a series of observed events $(h, r, t, \tau)$. At any time $\tau > \tau_0$, the TKG can be updated by the events before time $\tau$. The TPP can be characterized by conditional intensity function $\lambda (\tau)$. Given the historical events before a timestamp, if we can find a conditional intensity function to characterize them, then we can Predict whether the events will occur in the future with a conditional density and when the events will occur with an expectation. 

Know-Evolve~\citep{DBLP:conf/icml/TrivediDWS17-know-evolve} combines the TPP and the deep neural network framework to model the occurrence of facts as a multi-dimensional TPP. It characterizes the TPP with the Rayleigh process and uses neural networks to simulate the intensity function. RNN is used to learn the dynamic representation of the entities, and bilinear relationship score is used to capture multiple relational interactions between entities to modulate the intensity function. Thus, it can predict whether and when the event will occur.

GHNN~\citep{DBLP:conf/akbc/HanM0GT20-ghnn}  believes that the Hawkes process~\citep{hawkes1971spectra-hawkes-process}  based on the neural network can effectively capture the influence of past facts on future facts, and proposes a graph Hawkes neural network (GHNN). Firstly, it solves the problem that Know-Evolve could not deal with co-occurrence facts and uses a neighborhood aggregation module to process multiple facts of entities co-occurring. Then, it utilizes the continuous-time LSTM (cLSTM) model~\citep{DBLP:conf/nips/MeiE17-clstm}  to simulate the Hawkes process to capture the evolving dynamics of the facts to implement link prediction and time prediction.

EvoKG~\citep{10.1145/3488560.3498451-evokg}  argues that the above methods based on TPP lack to model the evolving network structure, and the methods based on autoregression lack to model the event time. It proposes a model jointly modeling the evolving network structure and event time. First, it uses an extended R-GCN and RNN to learn the time-evolving structural representations of entities and relations and utilizes an MLP with softmax to model the conditional probability of event triple. Then, it uses the same framework to learn the time-evolving temporal representations and adopts the TPP based on conditional density estimation with a mixture of log-normal distributions to model the event time. Finally, it jointly trains the two tasks and predicts the event and time in the future. 

\subsection{Interpretability-based Methods}
The aforementioned methods has resulted in a lack of interpretability and transparency in the generated results. As a result, we categorize interpretability-based methods as a separate category to underscore the crucial role of interpretability in developing reliable and transparent models. These methods aim to provide explanations for the predictions made by the models. Two popular types of such methods are subgraph reasoning-based and reinforcement learning-based approaches. 

\textbf{Subgraph Reasoning} 
xERTE~\citep{DBLP:conf/iclr/HanCMT21-xerte} is a subgraph reasoning-based method that proposes an explainable reasoning framework for predicting facts in the future. It starts from the head entity in the query and utilizes a temporal relational graph attention mechanism to learn the entity representation and relation representation. Then it samples the edges and temporal neighbors iteratively to construct the subgraph after several rounds of expansion and pruning. Finally, it predicts the tail entity in the subgraph. 

\textbf{Reinforcement Learning} 
Reinforcement learning (RL)~\citep{DBLP:journals/jair/KaelblingLM96-rl} is usually modeled as a Markov Decision Process (MDP)~\citep{Bel-markov}, which includes a specific environment and an agent. The agent has an initial state. After performing an action, it receives a reward from the environment and transitions to a new state. The goal of reinforcement learning is to find the policy network to obtain the maximum reward from all action strategies. 

CluSTeR~\citep{DBLP:conf/acl/LiJGLGWC20-cluster} proposes a two-stage reasoning strategy to predict the facts in the future. First, the clue related to a given query is searched and deduced from the history based on reinforcement learning. Then, the clue at different timestamps is regarded as a subgraph related to the query, and the R-GCN and GRU are used to learn the evolving representations of entities in the subgraph. Finally, the two stages are jointly trained, and the prediction is inferred. 

TITer~\citep{DBLP:conf/emnlp/SunZMH021-titer} directly uses the temporal path-based reinforcement learning model to learn the representations of the TKG and reasons for future facts. It adds temporal edges to connect each historical snapshot of the TKG. The agent starts from the head entity of the query, transitions to the new node according to the policy network, and searches for the answer node. The method designs a time-shaped reward based on Dirichlet distribution to guide the model learning. 

\subsection{Language Model}
In the domain of TKG, the rapid development of language models has prompted researchers to explore their application for predictive tasks. The current methodologies employing language models in the TKG domain predominantly encompass two distinct approaches: In-Context Learning and Supervised Fine-Tuning.

\textbf{In-Context Learning} ICLTKG\citep{lee2023temporal} introduces a novel TKG forecasting approach that leverages large language models (LLMs) through in-context learning (ICL)\citep{brown2020language} to efficiently capture and utilize irregular patterns of historical facts for accurate predictions. The implementation algorithm of this paper involves a three-stage pipeline designed to harness the capabilities of large language models (LLMs) for temporal knowledge graph (TKG) forecasting. The first stage focuses on selecting relevant historical facts from the TKG based on the prediction query. These facts are then used as context for the LLM, enabling it to capture temporal patterns and relationships between entities. Then the contextual facts are transformed into a lexical prompt that represents the prediction task. Finally, the output of the LLM is decoded into a probability distribution over the entities within the TKG. Throughout this pipeline, the algorithm controls the selection of background knowledge, the prompting strategy, and the decoding process to ensure accurate and efficient TKG forecasting. By leveraging the capabilities of LLMs and harnessing the irregular patterns embedded within historical data, this approach achieves competitive performance across a diverse range of TKG benchmarks without the need for extensive supervised training or specialized architectures.

zrLLM\citep{ding2023zero} introduces a novel approach that leverages large language models (LLMs) to enhance zero-shot relational learning on temporal knowledge graphs (TKGs). It proposes a method to first use an LLM to generate enriched relation descriptions based on textual descriptions of KG relations, and then a second LLM is employed to generate relation representations, which capture semantic information. Additionally, a relation history learner is developed to capture temporal patterns in relations, further enabling better reasoning over TKGs. The zrLLM approach is shown to significantly improve the performance of TKGF models in recognizing and forecasting facts with previously unseen zero-shot relations. Importantly, zrLLM achieves this without further fine-tuning of the LLMs, demonstrating the potential of alignment between the natural language space of LLMs and the embedding space of TKGF models. Experimental results exhibit substantial gains in zero-shot relational learning on TKGs, confirming the effectiveness and adaptability of the proposed zrLLM approach.

\textbf{Supervised Fine-Tuning} ECOLA~\citep{han2023ecola} proposes a joint learning model by leveraging text knowledge to enhance temporal knowledge graph representations. As existing TKGs often lack fact description information, the authors construct three new datasets that contain such information. During model training, they jointly optimize the knowledge-text prediction (KTP) objective and the temporal knowledge embedding (tKE) objective to improve the representation of TKGs. KTP employs pre-trained language models such as transformers~\citep{NIPS2017_3f5ee243-transformer}, while tKE can utilize an existing TKGRL model such as DyERNIE\citep{han2020dyernie}. By augmenting the temporal knowledge graph representation with text descriptions, the model achieves significant performance gains.

Frameworks such as GenTKG~\citep{liao2023gentkg} and Chain of History~\citep{luo2024chain} adopt retrieval augmented generation method for prediction. They utilize specific strategies to retrieve historical facts with high temporal relevance and logical coherence. Subsequently, these frameworks apply supervised fine-tuning language models to predict the future based on the retrieved historical facts. The input of the language model comprises historical facts and prediction queries, with the model outputting forecasted results. The authors have constructed a bespoke dataset of instructional data, which is utilized to train the language model, resulting in exemplary performance.

\subsection{Few-shot Learning Methods}
Few-shot learning (FSL)~\citep{10.1145/3386252-fewshotsurvey} is a type of machine learning problems that deals with the problem of learning new concepts or tasks from only a few examples. FSL has applications in a variety of domains, including computer vision~\citep{DBLP:conf/cvpr/KangC22-asnet}, natural language processing~\citep{9319339-fewshotNLPsurvey}, and robotics~\citep{DBLP:journals/ral/LiuZS23-fewshotRobotic}, where data may be scarce or expensive to acquire.
In TKGs, some entities and relations are only exist in a limited number of facts, and new entities and relations emerge over time. The latest TKGRL models now have the ability to perform FSL~\citep{10144560-fewshotkgsurvey}, which is essential for better representing these limited data. Due to the differences in handling data and learning methods for few entities and few relations, we will introduce them separately.

\textbf{Few Entities}
MetaTKG~\citep{xia-etal-2022-metatkg} reveals that new entities emerge over time in TKGs, and appear in only a few facts. Consequently, learning their representation from limited historical information leads to poor performance. Therefore, the authors propose a temporal meta-learning framework to address this issue.  Specifically, they first divide the TKG into multiple temporal meta-tasks, then employ a temporal meta-learner to learn evolving meta-knowledge across these tasks, finally, the learned meta-knowledge guides the backbone (which can be an existing TKGRL model, such as RE-GCN~\citep{10.1145/3404835.3462963-re-gcn}) to adapt to new data.
 
MetaTKGR~\citep{wang2022learning-MetaTKGR} confirms that emerging entities, which exist in only a small number of facts, are insufficient to learn their representations using existing models. To address this issue, the authors propose a meta temporal knowledge graph reasonging framework. The model leverages the temporal supervision signal of future facts as feedback to dynamically adjust the sampling and aggregation neighborhood strategy, and encoder the new entity representations. The optimized parameters can be learned via a bi-level optimization, where inner optimization initializes entity-specific parameters using the global parameters and fine-tunes them on the support set, while outer optimization operates on the query set using a temporal adaptation regularizer to stabilize meta temporal reasoning over time. The learned parameters can be easily adapted to new entities.

As existing datasets themselves often contain new entities that are associated with only a few facts, it is possible to directly divide the existing dataset into tasks to construct the support set and query set without requiring the creation of a new dataset. 

\textbf{Few Relations}
TR-Match~\citep{10.1007/978-3-031-30672-3_52-trmatch} identifies that most relations in TKGs have only a few quadruples, and new relations are added over time. Existing models are inadequate in addressing the few-shot scenario and may not fully represent the evolving temporal and relational features of entities in TKGs. To address these issues, the authors propose a temporal-relational matching network. Specifically, the proposed approach incorporates a multi-scale time-relation attention encoder to adaptively capture local and global information based on time and relation to tackle the dynamic properties problem. A new matching processor is designed to address the few-shot problem by mapping the query to a few support quadruples in a relation-agnostic manner. To address the challenge posed by few relations in temporal knowledge graphs, three new datasets, namely ICEWS14-few, ICEWS05-15-few, and ICEWS18-few, are constructed based on existing TKG datasets. The proposed TR-Match framework is evaluated on these datasets, and the experimental results demonstrate its capability to achieve excellent performance in few-shot relation scenarios.

MTKGE~\citep{10.1145/3543507.3583279-MTKGE}  recognizes that TKGs are subject to the emergence of unseen entities and relations over time. To address this challenge, the authors propose a meta-learning-based temporal knowledge graph extrapolation model. The proposed approach includes a Relative Position Pattern Graph (RPPG) to construct several position patterns, a Temporal Sequence Pattern Graph (TSPG) to learn different temporal sequence patterns, and a Graph Convolutional Network (GCN) module for extrapolation. This model leverages meta-learning techniques to adapt to new data and extract useful information from the existing TKG. The proposed MTKGE framework represents an important advancement in TKGRL by introducing a novel approach to knowledge graph extrapolation.

\subsection{Other Methods}
Other methods leverage the unique characteristics of TKGs to learn entity, relation, and timestamp representations. For instance, One approach explores the repetitive patterns of TKG and learns a more expressive representation using the copy-generation patterns. Alternatively, other methods employ various geometric and algorithmic techniques to capture the structural properties of TKGs and learn effective representations.

\textbf{Copy Generation}  
CygNet~\citep{DBLP:conf/aaai/ZhuCFCZ21-cygnet} finds that many facts show repeated patterns along the timeline, which indicates that potential knowledge can be learned from historical facts. Therefore, it proposes a time-aware copy-generation model which can predict future facts concerning the known facts. It constructs a historical vocabulary with multi-hot vectors of head entity and relation in each snapshot. In copy mode, it generates an index vector with an MLP to obtain the probability of the tail entities in the historical entity vocabulary. In generation mode, it uses an MLP to get the probability of tail entities in the entire entity vocabulary. After combining the two probabilities, it predicts the tail entity.

\textbf{Neural Ordinary Equation}
TANGO~\citep{DBLP:conf/emnlp/HanDMGT21-tango} contends that existing approaches for modeling TKGs are inadequate in capturing the continuous evolution of TKGs over time due to their reliance on discrete state spaces. To address this limitation, TANGO proposes a novel model based on Neural Ordinary Differential Equations (NODEs).
More specifically, the proposed model first employs a multi-relational graph convolutional network module to capture the graph structure information at each point in time. A graph transformation module is also utilized to model changes in edge connectivity between entities and their neighbors. The output of these modules is integrated to obtain dynamic representations of entities and relations. Subsequently, an ODE solver is adopted to solve these dynamic representations, thereby enabling TANGO to learn continuous-time representations of TKGs. TANGO's novel approach based on ODEs offers a more effective and accurate method for modeling the dynamic evolution of TKGs compared to existing techniques that rely on discrete representation space.

\textbf{Geometric model}
The transformation models represent TKGs in a Euclidean space, while Dyernie~\citep{han2020dyernie} maps the TKGs to a non-Euclidean space and employs Riemannian manifold product to learn evolving entity representations. This approach can model multiple simultaneous non-Euclidean structures, such as hierarchical and cyclic structures, to more accurately capture the complex structural properties of TKGs. By leveraging these additional structural features, the Dyernie method can more effectively capture the relationships between entities in TKGs, resulting in improved performance on TKGRL tasks.

BoxTE~\citep{DBLP:conf/aaai/MessnerAC22-boxte} is an innovative model that builds on the work of BoxE~\citep{DBLP:conf/nips/AbboudCLS20-boxe} and proposes a flexible and powerful framework that is fully expressive and inductive. It represents entities using a formulation that combines a base position vector $\boldsymbol{h,t} \in \mathbb{R}^d$, a translational bump vector $\boldsymbol{b_h,b_t} \in \mathbb{R}^d$, and a time bump vector $\boldsymbol{\tau^r} \in \mathbb{R}^d$. Specifically, the entity representations are defined as $\boldsymbol{h^{r(h,t|\tau)} = h + b_t + \tau^r}$, $\boldsymbol{t^{r(h,t|\tau)} = t +}$ $ \boldsymbol{b_h + \tau^r}$. Furthermore, BoxTE represents boxes using a time-induced relation head box $\boldsymbol{r^{h|\tau}} = \boldsymbol{r^h-\tau^r}$ and a time-induced relation tail box $\boldsymbol{r^{t|\tau} = \boldsymbol{r^t}}$ $\boldsymbol{-\tau^r}$. To determine whether a fact $r(h,t|\tau)$ holds, BoxTE checks if the representations for $h$ and $t$ lie within their corresponding boxes. This is expressed mathematically as follows: $\boldsymbol{h^{r(h,t|\tau)}} \in \boldsymbol{r^h}$, and $\boldsymbol{t^{r(h,t|\tau)}} \in \boldsymbol{r^t}$. By substracting $\boldsymbol{\tau^r}$ from both sides, it obtains the equivalent expressions: $\boldsymbol{h^{r(h,t)}} \in \boldsymbol{r^{h|\tau}}$, $\boldsymbol{t^{r(h,t)}} \in \boldsymbol{r^{t|\tau}}$. BoxTE enables the model to capture rigid inference patterns and cross-time inference patterns, thereby making it a powerful tool for TKGRL.

TGeomE~\citep{9713947-TGEOME} moves beyond the use of complex or hypercomplex spaces for TKGRL and instead proposes a novel geometric algebra-based embedding approach. This method utilizes multivector representations and performs fourth-order tensor factorization of TKGs while also introducing a new linear temporal regularization for time representation learning. The proposed TGeomE approach can naturally model temporal relations and enhance the performance of TKGRL models.

\begin{table}
  \caption{A summary of recent TKGRL methods.}
  \label{tab:models}
  \centering
  \scalebox{0.5}{
  \begin{tabular}{llccc}
    \hline
    Category & Model & Representation space & Encoder & Decoder \\
    \hline
    Transformation 
      & TTransE(2018) & $\mathbb{R}^d$ 
      & - & $\Vert\boldsymbol{h}+\boldsymbol{r}+\boldsymbol{\tau}-\boldsymbol{t}\Vert_{1/2}$ \\
      & TA-TransE(2018) & $\mathbb{R}^d$  
      & $r_{seq} = LSTM(r:\tau)$ & $\Vert\boldsymbol{h}+\boldsymbol{r}_{seq}-\boldsymbol{t}\Vert_{2}$ \\
      & HyTE(2018) & $\mathbb{R}^d$ 
      & $\boldsymbol{h}_{\tau} = \boldsymbol{h}-(\boldsymbol{w}_{\tau}^T\boldsymbol{h})\boldsymbol{w}_{\tau}$ & $\Vert \boldsymbol{h}_{\tau} + \boldsymbol{r}_{\tau} -\boldsymbol{t}_{\tau} \Vert_{1/2}$ \\      
      & Tero(2020) & $\mathbb{C}^d$
      & $\boldsymbol{h}_{\tau} = \boldsymbol{h} \circ \boldsymbol{\tau}$ & $\Vert \boldsymbol{h}_{\tau}+\boldsymbol{r}-\overline{\boldsymbol{t}}_{\tau} \Vert$ \\
      & ChronoR(2021) & $\mathbb{R}^{d \times{k}}$ 
      & $\boldsymbol{h}_{r,\tau} = \boldsymbol{h} \circ \left[\boldsymbol{r}|\boldsymbol{\tau}\right] \circ \boldsymbol{r}_2$ & $\left\langle \boldsymbol{h}_{r,\tau}, \boldsymbol{t}\right\rangle$ \\
      & RotateQVS(2022) & $\mathbb{Q}^d$ 
      & $\boldsymbol{h}_{\tau} = \boldsymbol{\tau} \boldsymbol{h} \boldsymbol{\tau}^{-1} $ & $\Vert \boldsymbol{h}_{\tau}+\boldsymbol{r}-\overline{\boldsymbol{t}}_{\tau} \Vert$ \\     
    \hline
    Decomposition     
      & DE-SimplE(2020) & $\mathbb{R}^d$ 
      & $\boldsymbol{e}_{\tau}[n]=\left\{\begin{array}{ll} \boldsymbol{e}[n]\sigma(\boldsymbol{w}[n]\tau+\boldsymbol{b}[n]),  & 1 \leq n \leq \gamma d  \\ \boldsymbol{e}[n],  & \gamma d \le n \leq d\end{array}\right.$ & $\frac{1}{2}(\left\langle \boldsymbol{h}_{\tau},\boldsymbol{r},\boldsymbol{t}_{\tau} \right\rangle + \left\langle \boldsymbol{t}_{\tau},\boldsymbol{r^{-1}},\boldsymbol{h}_{\tau} \right\rangle)$ \\   
      & T-SimplE(2020) & $\mathbb{R}^d$ 
      & - & $\frac{1}{2}(\left\langle \boldsymbol{h},\boldsymbol{r},\boldsymbol{t},\boldsymbol{\tau} \right\rangle + \left\langle \boldsymbol{t},\boldsymbol{r}^{-1},\boldsymbol{h},\boldsymbol{\tau} \right\rangle)$ \\
      & TComplEx(2020) & $\mathbb{C}^d$ 
      & - & $re(\left\langle \boldsymbol{h},\boldsymbol{r},\overline{\boldsymbol{t}},\boldsymbol{\tau} \right\rangle)$ \\   
      & TLT-KGE(2022) & $\mathbb{H}^d$ 
      & $\begin{array}{ll} \boldsymbol{h}^q_{\tau}=\boldsymbol{h_a}+\boldsymbol{h_bi}+\boldsymbol{\tau_{e,c}j}+\boldsymbol{\tau_{e,d}k}, \\ 
\boldsymbol{t}^q_{\tau}=\boldsymbol{t_a}+\boldsymbol{t_bi}+\boldsymbol{\tau_{e,c}j}+\boldsymbol{\tau_{e,d}k}, \\
\boldsymbol{r}^q_{\tau}=\boldsymbol{r_a}+\boldsymbol{r_bi}+\boldsymbol{\tau_{r,c}j}+\boldsymbol{\tau_{r,d}k} \end{array}$ & $\left<\boldsymbol{h}^q_{\tau} \odot \boldsymbol{r}^q_{\tau},\boldsymbol{t}^q_{\tau}\right> + \left<(\boldsymbol{h}^q_{\tau} \odot \boldsymbol{r}^q_{\tau})',\boldsymbol{t}^q_{\tau}\right>$\\ 
      & TuckERT(2022) & $\mathbb{R}^d$ 
      & - & $\left\langle \boldsymbol{\mathcal{M}}; \boldsymbol{h},\boldsymbol{r},\boldsymbol{t},\boldsymbol{\tau} \right\rangle$ \\
    \hline
    Graph Neural Networks
     & TEA-GNN(2021) & $\mathbb{R}^d$ 
     & \emph{Time-aware GNN} & - \\
     & TREA(2022) & $\mathbb{R}^d$ 
     & \emph{Temporal Relational GAT} & - \\
     & DEGAT(2022) & $\mathbb{R}^d$ 
     & \emph{GAT} & \emph{DE-ConvKB} \\  
     & $L^2$TKG(2023) & $\mathbb{R}^d$ 
     & \emph{R-GCN + GRU} & \emph{ConvTransE} \\  
    \hline
    Capsule Network
     & TempCaps(2022) & $\mathbb{R}^d$  
     & \emph{Temporal Capsule Network} & \emph{MLP} \\
     & BiQCap(2023) & $\mathbb{H}^d$  
     & \emph{Biquaternion + Capsule Network} & \emph{MLP} \\
     & DuCape(2023) & $\mathbb{H}^d$  
     & \emph{Dual Quaternion + Capsule Network} & \emph{MLP} \\
    \hline
    Autoregression
     & RE-NET(2020) & $\mathbb{R}^d$  
     & \emph{R-GCN + GRU} & \emph{MLP} \\
     & Glean(2020) & $\mathbb{R}^d$ 
     & \emph{(CompGNC + GCN) + RNN} & - \\
     & RE-GCN(2021) & $\mathbb{R}^d$ 
     & \emph{relation-aware GCN + GRU} & \emph{ConvTransE} \\
     & TiRGN(2022) & $\mathbb{R}^d$ 
     & \emph{relation-aware GCN + GRU} & \emph{Time-ConvTransE} \\
     & Cen(2022) & $\mathbb{R}^d$ 
     & \emph{relation-aware GCN + GRU} & \emph{CNN} \\
    \hline
    TPP
     & Know-Evolve(2017) & $\mathbb{R}^d$ 
     & \emph{Rayleigh process} & - \\
     & GHNN(2020) & $\mathbb{R}^d$ 
     & \emph{Graph Hawkes Process} & - \\
     & EvoKG(2022) & $\mathbb{R}^d$ 
     & \emph{Structure module + Temporal module} & - \\
    \hline
    Interpretability
     & xERTE(2021) & $\mathbb{R}^d$ 
     & \emph{TRGA} & - \\
     & CluSTeR(2021) & $\mathbb{R}^d$ 
     & \emph{RL + R-GCN + GRU} & - \\
     & TiTer(2021) & $\mathbb{R}^d$ 
     & \emph{RL} & - \\
    \hline
    Language model
     & ICLTKG(2023) & $\mathbb{R}^d$ 
     & - & \emph{Transformer} \\
     & zrLLM(2023) & $\mathbb{R}^d$ 
     & - & \emph{Transformer} \\
     & ECOLA(2022) & $\mathbb{R}^d$ 
     & - & \emph{Transformer + Dyernie} \\ 
     & GENTKG(2023) & $\mathbb{R}^d$ 
     & - & \emph{Transformer} \\
     & Chain of History(2024) & $\mathbb{R}^d$ 
     & - & \emph{Transformer} \\
    \hline
    Few-shot Learning
     & MetaTKG(2022) & $\mathbb{R}^d$ 
     & \emph{REGCN} & - \\
     & MetaTKGR(2022) & $\mathbb{R}^d$ 
     & \emph{Time-aware GAT} & - \\
     & TR-Match(2023) & $\mathbb{R}^d$ 
     & \emph{Time-Relation Attention} & - \\
     & MTKGE(2023) & $\mathbb{R}^d$ 
     & \emph{RPPG + TSPG + GCN} & - \\
    \hline    
    Others     
     & CygNet(2021) & $\mathbb{R}^d$ 
     & \emph{Copy mode+ Generation mode} & - \\
     & TANGO(2021) & $\mathbb{R}^d$ 
     & \emph{MGCN + Trans + ODE Solver} & Distmult or TuckER \\
     & Dyernie(2020) & $\mathbb{R}^d$ 
     & $\boldsymbol{e_\tau^{(i)}} = exp_0^{K_i}(log_0^{K_i}(\boldsymbol{\overline{e}^{(i)})+v_{e^{(i)}})}$ & $\begin{array}{ll}\sum_{i=1}^k-d_{M_{K_i}^{n_i}}(\boldsymbol{P^{(i)}} \otimes_{K_i} \boldsymbol{h_\tau^{(i)}}, \\ \boldsymbol{t_\tau^{(i)}} \oplus_{K_i} \boldsymbol{p^{(i)}}) + b_h^{(i)} + b_t^{(i)} \end{array}$ \\
     & BoxTE(2022) & $\mathbb{R}^d$ 
     & $\begin{array}{ll}\boldsymbol{h^{r(h,t)} = h + b_t},\\ \boldsymbol{t^{r(h,t)} = t + b_h},\\ \boldsymbol{r^{h|\tau}} = \boldsymbol{r^h-\tau^r},\\ \boldsymbol{r^{t|\tau}} = \boldsymbol{r^t-\tau^r}\end{array}$ 
     & $\Vert d(\boldsymbol{h^{r(h,t)}}, \boldsymbol{r^{h|\tau}})\Vert + \Vert d(\boldsymbol{t^{r(h,t)}}, \boldsymbol{r^{t|\tau}})\Vert$ \\
     & TGeomE(2023) & $\mathbb{G}^{n\times d}$ 
     & $\boldsymbol{h,r,t} = [M_1,...,M_k]$ & $\left\langle Sc(\boldsymbol{h \otimes_n r \otimes_n \overline{t}}),1\right\rangle$ \\     
  \hline
\end{tabular}
}
\end{table}

\subsection{Summary} 
In this section, we divide the TKGRL methods into ten categories and introduce the core technologies of these methods in detail. Table~\ref{tab:models} shows the summary of the methods, including the representation space, the encoder for mapping the entities and relations to the vector space, and the decoder for predicting the answer.

\section{Applications of Temporal Knowledge Graph}
By introducing temporal information, TKG can express the facts in the real world more accurately, improve the quality of knowledge graph representation learning, and answer temporal questions more reliably. It is helpful for the applications such as reasoning, entity alignment, and question answering.

\subsection{Temporal Knowledge Graph Reasoning}
TKGRL methods are widely used in temporal knowledge graph reasoning (TKGR) tasks which automatically infers new facts by learning the existing facts in the KG. TKGR usually has three subtasks: entity prediction, relation prediction, and time prediction. Entity prediction is the basic task of link prediction, which can be expressed as two queries $(?, r, t, \tau)$ and $(h, r, ?, \tau)$. Relation prediction and time prediction can be expressed as $(h, ?, t, \tau)$ and $(h, r, t, ?)$, respectively.

TKGR can be divided into two categories based on when the predictions of facts occur, namely interpolation and extrapolation. Suppose that a TKG is available from time $\tau_0$ to $\tau_T$. The primary objective of interpolation is to retrieve the missing facts at a specific point in time $\tau \ (\tau_0 \leq \tau \leq \tau_T)$. This process is also known as temporal knowledge graph completion (TKGC). On the other hand, extrapolation aims to predict the facts that will occur in the future $(\tau \geq \tau_T)$ and is referred to as temporal knowledge graph forecasting.

Several methods have been proposed for Temporal Knowledge Graph Completion (TKGC) including transformation-based, decomposition-based, graph neural networks-based, capsule Network-based, and other geometric methods. These techniques aim to address the problem of missing facts in TKGs by leveraging various mathematical models and neural networks. In contrast, predicting future facts in TKGs requires a different approach that can model the temporal evolution of the graph. Autoregression-based, temporal point process-based, and few-shot learning methods are commonly used for this task. Interpretability-based methods are used to increase the reliability of prediction results. These techniques provide evidence to support predictions, helping to establish trust and improving the overall quality of predictions made by the model. To further enhance the performance of TKGRL, semantic augmentation technology can be employed to improve the quality and quantity of semantic information of TKGs. Utilizing entity and relation names, as well as textual descriptions of fact associations, can enrich their representation and promote the development of downstream tasks of TKGs. In addition, large language models (LLMs) for natural language processing (NLP) can facilitate the acquisition of rich semantic information about entities and relations, further augmenting the performance of TKGRL models.

\subsection{Entity Alignment Between Temporal Knowledge Graphs}
Entity alignment (EA) aims to find equivalent entities between different KGs, which is important to promote the knowledge fusion between multi-source and multi-lingual KGs. Defining $G_1=(E_1,R_1,T_1,F_1)$ and $G_2=(E_2,R_2,T_2,F_2)$ to be two TKGs, $S=\{(e_{1_i},e_{2_j})|e_{1_i}\in E_1,e_{2_j}\in E_2\}$ is the set of alignment seeds between $G_1$ and $G_2$. EA seeks to find new alignment entities according to the alignment seeds $S$. The methods of EA between TKGS mainly adopt the GNN-based model.

Currently, exploring the entity alignment (EA) between Temporal Knowledge Graphs (TKGs) is an active area of research. TEA-GNN~\citep{xu-etal-2021-time-tea-gnn} was the first method to incorporate temporal information via a time-aware attention Graph Neural Network (GNN) to enhance EA.
TREA~\citep{10.1145/3485447.3511922-TREA} utilizes a temporal relational attention GNN to integrate relational and temporal features of entities for improved EA performance.
STEA~\citep{cai-etal-2022-stea} identifies that the timestamps in many TKGs are uniform and proposes a simple GNN-based model with a temporal information matching mechanism to enhance EA. Initially, the structure and relation features of an entity are fused together to generate the entity embedding. Then, the entity embedding is updated using GNN aggregation from neighborhood. Finally, the entity embedding is obtained by concatenating the embedding of each layer of the GNN. STEA not only updates the representation of entities but also calculates time similarity by considering associated timestamps. The method combines both the similarities of entity embeddings and the similarities of entity timestamps to obtain aligned entities. Overall, STEA offers an effective way of improving entity representation in TKGs and provides a reliable solution for aligning entities over time.

\subsection{Question Answering Over Temporal Knowledge Graphs}
Question answering over KG (KGQA) aims to answer natural language questions based on KG. The answer to the question is usually an entity in the KG. In order to answer the question, one-hop or multi-hop reasoning is required on the KG. Question answering over TKG (TKGQA) aims to answer temporal natural language questions based on TKG, the answer to the question is entity or timestamp in the TKG, and the reasoning on TKG is more complex than it on KG.

Research on TKGQA is in progress. CRONKGQA~\citep{DBLP:conf/acl/SaxenaCT20-cronkgqa} release a new dataset named CRONQUESTIONS and propose a model combining representation of TKG and question for TKGQA. It first uses TComplEx to obtain the representation of entities and timestamps in the TKG, and utilizes BERT~\citep{DBLP:conf/naacl/DevlinCLT19-bert} to obtain their representations in the question, then calculates the scores of all entities and times, and finally concatenated the score vectors to obtain the answer.

TSQA~\citep{DBLP:conf/acl/ShangW0022-tsqa} argues existing TKGQA methods haven't explore the implicit temporal feature in TKGs and temporal questions. It proposes a time sensitive question answering model which consists of a time-aware TKG encoder and a time-sensitive question answering module. The time-aware TKG encoder uses TComplEx with time-order constraints to obtain the representations of entities and timestamps. The time-sensitive question answering module first decomposes the question into entities and a temporal expression. It uses the entities to extract the neighbor graph to reduce the search space of timestamps and answer entities. The temporal expression is fed into the BERT to learn the temporal question representations. Finally, entity and temporal question representations are combined to estimate the time and predict the entity with contrastive learning.

\section{Future Directions}
Despite the significant progress made in TKGRL research, there remain several important future directions. These include addressing scalability challenges, improving interpretability, incorporating information from multiple modalities, and leveraging large language models to enhance the ability of representing dynamic and evolving TKGs. 

\subsection{Scalability} 
The current datasets available for TKG are insufficient in size compared to real-world knowledge graphs. Moreover, TKGRL methods tend to prioritize improving task-specific performance and often overlook the issue of scalability. Therefore, there is a pressing need for research on effective methods of learning TKG representations that can accommodate the growing demand for data. A possible avenue for future research in this field is to investigate various strategies for enhancing the scalability of TKGRL models. 

One approach for improving the scalability of TKGRL models is to employ distributed computing techniques, such as parallel processing or distributed training, to enable more efficient processing of large-scale knowledge graphs. Parallel processing involves partitioning the dataset into smaller subsets and processing each subset simultaneously. In contrast, distributed training trains the model on various machines concurrently, with the outcomes combined to enhance the overall accuracy of the model. This approach could prove especially beneficial for real-time processing of extensive knowledge graphs in applications that require quick response times.

Another approach is to use sampling techniques to reduce the size of the knowledge graph without sacrificing accuracy. For example, researchers could use clustering algorithms to identify groups of entities and events that are highly interconnected and then sample a representative subset of these groups for training the model. This approach could help to reduce the computational complexity of the model without sacrificing accuracy. Sampling techniques can also be used for negative sampling in TKGRL. Negative sampling involves selecting negative samples that are not present in the knowledge graph to balance out the positive samples during training. By employing efficient negative sampling techniques, researchers can significantly reduce the computational complexity of the TKGRL model while maintaining high accuracy levels. 

Overall, addressing issues related to scalability will be critical for advancing the state-of-the-art in temporal knowledge graph research and enabling practical applications in real-world scenarios.

\subsection{Interpretability} 
The enhancement of interpretability is a crucial research direction, as it allows for better understanding of how model outputs are derived and ensures the reliability and applicability of the model's results. Despite the availability of interpretable methods, developing more interpretable models and techniques for temporal knowledge graphs remains a vital research direction. 

One promising approach involves incorporating attention mechanisms to identify the most relevant entities and events in the knowledge graph at different points in time. This approach would allow users to understand which parts of the graph are most important for a given prediction, which could improve the interpretability of the model.

In addition, researchers could explore the use of visualization techniques to help users understand the structure and evolution of the knowledge graph over time. For example, interactive visualizations could enable users to explore the graph and understand how different entities and events are connected.

By making TKGRL more interpretable, we can gain greater insights into complex real-world phenomena, support decision-making processes, and ensure that these models are useful for practical applications.

\subsection{Information Fusion} 
Most TKGRL methods only utilize the structural information of TKGs, with few models incorporating textual information of entities and relations. However, text data contains rich features that can be leveraged to enhance TKGs' representation. Therefore, effectively fusing various features of TKGs, including entity feature, relation feature, time feature, structure feature and textual feature, represents a promising future research direction.

One approach to information fusion in TKGRL is to use multi-modal data sources. For example, researchers can combine textual data, such as news articles or social media posts, with structured data from knowledge graphs to improve the accuracy of the model. This approach can help the TKGRL model to capture more relationships between entities and events that may not be apparent from structured data alone.

Another approach is to use attention mechanisms to dynamically weight the importance of different sources of information at different points in time. This approach would allow the model to focus on the most relevant information for a given prediction, which could improve the accuracy of the model while reducing computational complexity.

In general, information fusion is a powerful tool in TKGRL that can help researchers improve the accuracy and reliability of the model by combining information from multiple sources. However, it is essential to carefully weigh the benefits and costs of using different fusion techniques, depending on the specific dataset and research goals.

\subsection{Incorporating Large Language Models} 
Recent advances in natural language processing, such as the development of large language models (LLMs) ~\citep{zhao2023survey-llm} has been largely advanced by both academia and industry. A notable achievement in the field of LLMs is the introduction of ChatGPT~\footnote{https://openai.com/blog/chatgpt/}, a highly advanced AI chatbot. Developed using LLMs, ChatGPT has generated significant interest and attention from both the research community and society at large. ChatGPT uses the generative pre-trained transformer (GPT) such as GPT-4~\citep{openai2023gpt4}, have led to significant improvements in various natural language tasks. LLMs have been shown to be highly effective at capturing complex semantic relationships between words and phrases, and they may be able to provide valuable insights into the meaning and context of entities and relations in a knowledge graph. Efficiently combining LLMs with TKGRL is an novel research direction for the future.

One approach to incorporating LLMs into TKGRL is to use LLMs to generate embeddings for entities and relations. These embeddings could be used as input to a TKGRL model, enabling it to capture more rich feature of  entities and relations over time.

Another potential approach is to use LLMs to generate textual descriptions of entities and facts in the TKGs. These descriptions could be used to enrich the TKGs with additional semantic information, which could then be used to improve the accuracy of predictions.

Aboveall, incorporating LLMs into TKGRL has the potential to significantly improve the accuracy and effectiveness of these models, and it is an exciting area for future research. However, it is essential to carefully consider the challenges and limitations of using LLMs, such as computational complexity and potential bias in the pre-trained data.

\section{Conclusion}
Temporal knowledge graphs (TKGs) provide a powerful framework for representing and analyzing complex real-world phenomena that evolve over time. Temporal knowledge graph representation learning (TKGRL) is an active area of research that investigates methods for automatically extracting meaningful representations from TKGs.

In this paper, we provide a detailed overview of the development, methods, and applications of TKGRL. We begin by the definition of TKGRL and discussing the datasets and evaluation metrics commonly used in this field. Next, we categorize TKGRL methods based on their core technologies and analyze the fundamental ideas and techniques employed by each category. Furthermore, we offer a comprehensive review of various applications of TKGRL, followed by a discussion of future directions for research in this area. By focusing on these areas, we can continue to drive advancements in TKGRL and enable practical applications in real-world scenarios.




\section*{Acknowledgements}
We appreciate the support from National Natural Science Foundation of China with the Main Research Project on Machine Behavior and Human-Machine Collaborated Decision Making Methodology (72192820 \& 72192824), Pudong New Area Science \& Technology Development Fund (PKX2021-R05), Science and Technology Commission of Shanghai Municipality (22DZ-2229004), and Shanghai Trusted Industry Internet Software Collaborative Innovation Center.



\bibliographystyle{elsarticle-harv} 
\bibliography{TKGRL.bib}






\end{document}